\documentclass[letterpaper]{article}
\usepackage[preprint]{style}
\usepackage{natbib}
\usepackage[hyphens]{url}
\usepackage{graphicx}

\usepackage{booktabs}

\usepackage{amsmath}
\usepackage{amssymb}
\usepackage{makecell}
\usepackage{multirow}

\title{Structure-Aware Skill Retrieval and Arbitration for Optimization Modeling}
\author{Anran Liu, Benzhao Hou, Zongsheng Cao, Daichao Wang, Lang Chen, and Shitong Chen\corresponding}
\affiliations{}

\begin{document}

\maketitle

\begin{abstract}
Large Language Models (LLMs) can turn natural-language optimization tasks into solver-checked answers, but current agents remain brittle under narrative rewrites and often store reusable experience at the instance level. We introduce \textsc{ArchOR}, an archetype-centric skill agent that builds on \textsc{OptSkills} by adding two retrieval-focused modules. \emph{Centroid-Aware Representation Learning} (CARL) refines the ordering of cached skill representations through a low-rank prototype residual. \emph{Signature-Guided Retrieval} (S\textsuperscript{3}R) attaches open-dimensional structural signatures to skills, prioritizes candidates by weighted Jaccard overlap, and bounds semantic re-ranking to a top-$K$ set. A baseline-preserving arbitration rule determines when the structural route should replace the original selector. Experiments on the same five optimization benchmarks with the same DeepSeek-V4.1-Flash backbone and the unchanged released skill library show that these access-level changes produce consistent improvements in solver-verified accuracy.
\end{abstract}

\section{Introduction}
Operational optimization underlies many industrial decision pipelines, including routing, production scheduling, energy dispatch, and workforce planning. Turning these business descriptions into executable models is still a labor-intensive expert task: practitioners must identify decision variables, formalize constraints, choose objectives, and translate the formulation into solver code. Large language models (LLMs) have therefore become an attractive interface for optimization modeling. Recent systems improve this interface through multi-agent orchestration~\cite{pmlr-v235-ahmaditeshnizi24a, xiao2024chainofexperts}, domain-oriented fine-tuning~\cite{JiangShu2025llmopt, Huang2025}, or reinforcement learning with solver feedback~\cite{chen2025solverinformed, ding2026or}.

However, strong benchmark accuracy does not by itself solve the reuse problem that matters in practice. A deployed modeling agent should recognize that two differently narrated requests may instantiate the same optimization archetype, route repeated instances through a stable workflow, and extend its library when the distribution shifts. Recent experience-based systems move in this direction: AlphaOPT~\cite{kong2025alphaopt} maintains solver-verified insights, while LEAN-LLM-OPT~\cite{liang2026largescaleoptimizationmodelautoformulation} retrieves structured exemplars to seed workflow construction. Their stored artifacts are still largely instance-centered, so surface narratives, noisy descriptions, and missing coverage can distort retrieval even when the underlying mathematical structure is familiar.

The central difficulty is that optimization skills are defined by structure rather than by wording alone. Two assignment problems can mention different commodities, agents, and budgets while requiring the same variable pattern and constraint family; conversely, two descriptions can share terms such as ``distribute'' or ``mix'' while calling for incompatible formulations. A raw cached vector and a natural-language skill card have no explicit reason to preserve these archetype boundaries, leaving robust generalization and in-distribution routing unresolved. A convincing skill-reuse system should therefore show that its retrieval space and selection rules align with optimization structure.

We address this issue with \textsc{ArchOR}, an archetype-centric skill agent for LLM-based optimization modeling. Rather than modifying skill construction, \textsc{ArchOR} starts from the released \textsc{OptSkills} library, whose workflow-level skills consolidate successful and failed trajectories for optimization archetypes. The proposed system changes how a query is routed to those skills: CARL refines cached skill-vector ordering, S\textsuperscript{3}R imposes structural filtering, and baseline-preserving arbitration retains the original route when structural coverage is uncertain.
Two lightweight modules make this archetype view operational. \emph{Centroid-Aware Representation Learning} (CARL) transforms cached skill vectors with archetype prototypes and a low-rank residual so that neighboring skills are ordered by structural archetype rather than surface wording. \emph{Signature-Guided Retrieval} (S\textsuperscript{3}R) attaches an open-dimensional structural signature to every skill, with attributes such as \texttt{problem\_class}, \texttt{var\_pattern}, and \texttt{constraint\_family}. At test time, a query signature prioritizes compatible candidates by weighted overlap before LLM semantic re-ranking over a bounded candidate set. Because neither signal is reliable on every cross-domain query, \textsc{ArchOR} treats the inherited selector as a fallback and overrides it only when inference-time evidence supports the structural route.

Both components are needed because a fixed solver can be effective for one family of optimization structures yet fail when the variable pattern or constraint family changes. \textsc{ArchOR} therefore represents the problem structure, keeps a portfolio of complementary workflow skills, and routes each query to the most compatible workflow. CARL improves the ordering of cached archetype representations, while S\textsuperscript{3}R provides a discrete guardrail against semantically plausible but structurally wrong skills. Table~\ref{tab:knn_result} further validates the representation-level improvement on the learned skill library.

Empirically, with the same DeepSeek-V4.1-Flash backbone~\cite{deepseekai2026deepseekv41flash}, \textsc{ArchOR} produces consistent end-to-end accuracy improvements across five complementary benchmarks, while system-level and module ablations separate retrieval quality, routing behavior, and arbitration safety. Our contributions are threefold:
\begin{enumerate}
    \setlength{\itemsep}{0pt}
    \setlength{\parsep}{0pt}
    \setlength{\topsep}{0pt}
\item We identify retrieval as the limiting component of archetype-centric skill libraries: raw cached skill vectors can blur archetype boundaries, while semantics-only skill cards can make structurally incompatible workflows appear applicable.
\item We introduce two complementary modules that operationalize this view: one sharpens cached-vector neighborhoods, and the other records auditable structural attributes and uses their overlap to restrict the candidate set. A baseline-preserving arbitration rule prevents incomplete signature coverage from discarding reliable baseline skills.
\item We evaluate \textsc{ArchOR} across five benchmarks with the same backbone and released skill library; system-level ablations, module ablations, and leave-one-out retrieval diagnostics show when structural routing helps and when it must fall back to the baseline.
\end{enumerate}

\section{Related Work}
\label{relatedwork}
\subsection{Optimization Modeling with LLMs}

\paragraph{Post-Training Methods.} A first line of work embeds optimization competence directly into the model weights. Recent representatives cover supervised fine-tuning (ORLM~\cite{Huang2025}, OptMATH~\cite{lu2025optmathscalablebidirectionaldata}, AutoOR~\cite{motwani2026autoor}), preference optimization (LLMOPT~\cite{JiangShu2025llmopt}), and reinforcement learning with solver-based reward (SIRL~\cite{chen2025solverinformed}, OR-R1~\cite{ding2026or}). Solver feedback gives these models a stronger training signal than ordinary instruction tuning, especially when the objective value can be checked exactly. The cost is specificity: competence is tied to model parameters, updating it requires additional training, and a deployed system cannot easily inspect which learned procedure is being reused for a new query. \textsc{ArchOR} instead keeps the backbone frozen and places reusable procedures in an auditable skill library, so a routing decision can be explained by cached-vector similarity, signature overlap, and arbitration status.

These two families also imply different update cycles. A fine-tuned checkpoint can enforce a consistent modeling style and amortize competence into parameters, but correcting a recurrent formulation error may require another training run or a prompt workaround. A library can be versioned, inspected, and revised one workflow at a time. That transparency shifts the burden from learning a procedure to deciding when the procedure applies: an external skill is useful only if the retriever preserves its boundary conditions and the router does not override a safer known workflow.

\paragraph{Agent-Based Methods.} A second line keeps the base model frozen and instead composes a multi-stage LLM pipeline in which formulation, coding, execution, and repair are handled by distinct agents. Chain-of-Experts~\cite{xiao2024chainofexperts} and OptiMUS~\cite{pmlr-v235-ahmaditeshnizi24a} route sub-tasks through role-specialized experts glued together by a conductor. ORMind~\cite{wang2025ormind} and ORThought~\cite{yang2025orthought} layer cognitive-style heuristics on top of the pipeline so that business intent is progressively refined into a mathematical formulation and executable solver code. These systems improve error recovery and task decomposition, but their reusable knowledge is usually encoded as prompts, agent roles, or execution habits rather than as an explicitly indexed optimization workflow. Consequently, two problems with the same variable pattern may still be solved from scratch unless the pipeline itself maintains an external memory. \textsc{ArchOR} is complementary to this direction: it supplies a structure-aware retrieval layer that can decide when an existing workflow is applicable and when the original selector should remain authoritative.

\paragraph{Experience-Enhanced Methods.} The design question here is not merely whether experience should be reused, but which artifact is indexed and what boundary controls reuse. Constraint-centered systems such as DRoC~\cite{jiang2025droc} and ConstraintLLM~\cite{shi-etal-2025-constraintllm} are effective when the bottleneck is translating stated requirements into formal atoms, but isolated constraints do not specify the order of modeling decisions or the checking steps required after a solver returns. Exemplar-based systems such as MIRROR~\cite{shi2026mirror} and LEAN-LLM-OPT~\cite{liang2026largescaleoptimizationmodelautoformulation} provide fuller scaffolds, yet a retrieved instance can also import incidental quantities, objective conventions, or domain assumptions. AlphaOPT~\cite{kong2025alphaopt} makes those boundaries more explicit by refining insights from failed attempts, although a lesson must still be matched at the right level of generality. OptSkills~\cite{yang2026optskills} occupies an informative middle point: its reusable artifact is an end-to-end workflow, while archetype clustering reduces dependence on narrative details during library construction. \textsc{ArchOR} starts after that construction step. It does not change how workflows are formed; instead, it determines whether a stored workflow is admissible for the current query and whether it is safe to supersede the inherited baseline route.

A closely related concurrent system, OptiSkill~\cite{zhao2026optiskill}, also builds a reusable optimization-specific skill bank, organizing solver-verified experience into global strategies and step-level error-prevention rules and evolving the bank with validated test-time experience. Its emphasis is library construction and evolution. \textsc{ArchOR} instead keeps the released \textsc{OptSkills} library frozen and studies the downstream access decision: candidate admissibility, ordering, and override safety.

\subsection{From Skill Acquisition to Skill Access}
Research on agent skills can be viewed around a lifecycle: XSkill~\cite{jiang2026xskill} asks what part of an execution trace should be abstracted into transferable procedure; EvoSkill~\cite{alzubi2026evoskill}, Memento-Skills~\cite{zhou2026mementoskills}, and AutoSkill~\cite{yang2026autoskill} study how a catalog is discovered, routed to, and revised; Trace2Skill~\cite{ni2026trace2skill} consolidates local trajectory lessons; and MemSkill~\cite{zhang2026memskill} turns memory management itself into a learned competence. SkillRet~\cite{kang2026skillret} further shows that selecting one skill from a large library is itself a retrieval problem rather than an incidental prompt detail. This line shows that reusable knowledge must be neither an unstructured trace nor an overly broad prompt. It also makes clear that skill content and skill access are separate problems: a well-formed catalog can still fail if incompatible entries dominate the candidate list.

Optimization sharpens the access problem because compatibility is typed rather than merely topical. A candidate must agree on decision roles, conservation and capacity relations, objective orientation, integrality, and the assumptions expected by its solver backend. Two workflows can share a transportation vocabulary while imposing different directed-arc or shipment semantics; both may execute, but only one represents the stated decision problem. Multi-signal structural-semantic consistency has also been used to select generated MILP programs without solving every candidate~\cite{zeng2026milp}; \textsc{ArchOR} addresses the complementary earlier decision of retrieving an applicable workflow from a reusable skill library and deciding whether it may override a stable baseline.

This separation suggests three operational requirements for an access layer. The first is recall: plausible archetypes must remain visible after caching, filtering, and re-ranking. The second is typed admissibility: topical similarity cannot authorize a workflow whose decision roles or solver assumptions conflict with the query. The third is override safety: a new route should replace a stable baseline only when inference-time evidence indicates that it is executable and structurally credible. Treating these requirements separately avoids collapsing retrieval quality into end-to-end correctness, since a semantically near workflow can execute successfully yet encode the wrong decision semantics. The three components of \textsc{ArchOR} follow this decomposition: CARL improves candidate ordering, S\textsuperscript{3}R supplies explicit compatibility evidence, and arbitration decides whether that evidence is strong enough to supersede the inherited route.

\textsc{ArchOR} therefore contributes an access layer for a stable optimization-skill library. CARL improves the geometry of cached candidates, S\textsuperscript{3}R supplies explicit compatibility attributes, and arbitration controls when the resulting route may supersede the baseline. This separation is diagnostic as well as empirical: signature overlap, candidate rank, the chosen route, and solver validity distinguish omitted ingredients, over-broad aliases, ranking failures, and unsafe overrides. It also explains why nearest-neighbor gains need not imply end-to-end gains—an executable model with the wrong semantics can still return a finite objective.

\section{Methodology}
\label{methodology}

\begin{figure*}[t]
    \centering
    \includegraphics[width=\textwidth]{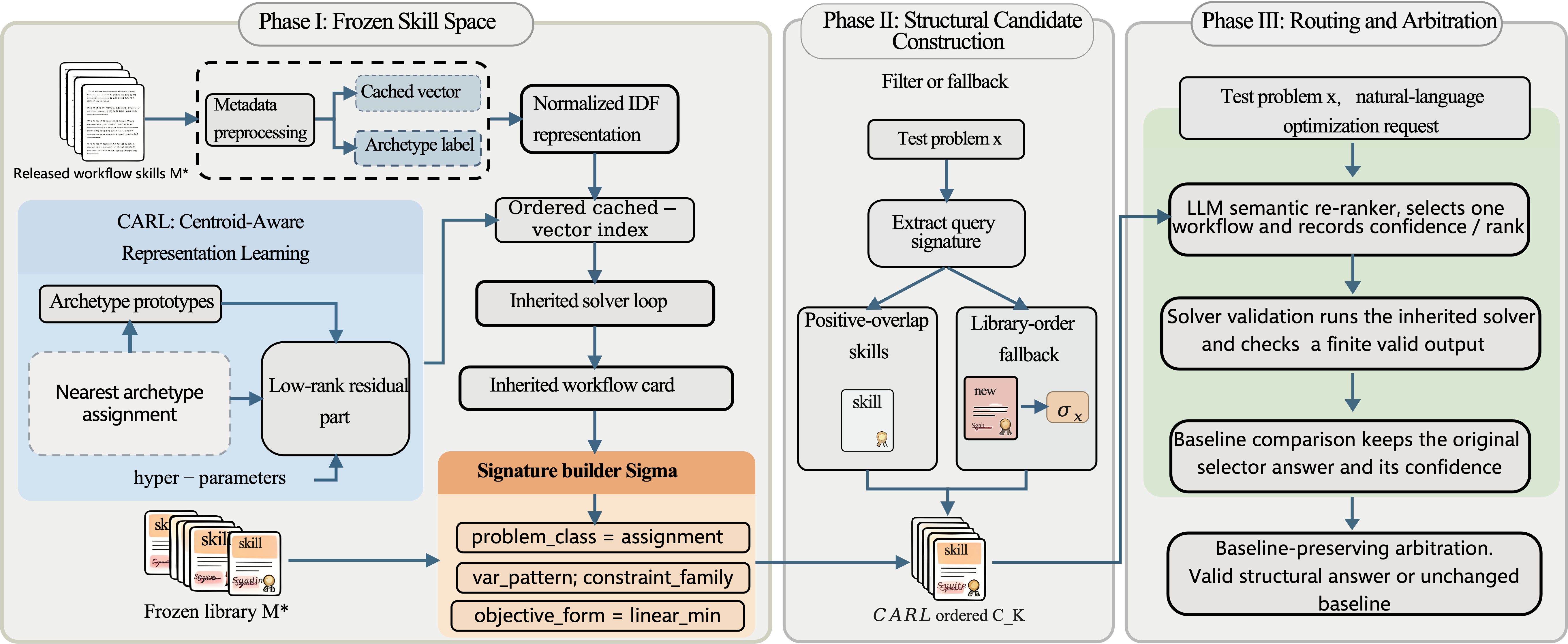}
\caption{
\textbf{Overview of \textsc{ArchOR}.} Starting from the \textsc{OptSkills} skill library, CARL orders cached skill representations by archetype similarity. Each skill and query also receives an open-dimensional structural signature. S$^3$R first filters skills by signature overlap and then invokes LLM semantic re-ranking only on the remaining candidates. Baseline-preserving arbitration decides whether to use the structural route or preserve the original \textsc{OptSkills} route.
}
\label{fig:framework}
\end{figure*}

\textsc{ArchOR} inherits the \textsc{OptSkills}~\cite{yang2026optskills} skill library, structural extractor, solver execution loop, and verification rule unchanged. Figure~\ref{fig:framework} summarizes the resulting inference pipeline. In the inherited interface, extracted optimization ingredients and a scenario-neutral rewrite are embedded and fused, while an LLM selector compares the query with natural-language skill metadata over the library. We keep this semantics-only route as the fallback baseline, but focus on the missing retrieval decision: which skills are structurally admissible, how those candidates should be ordered, and when the resulting structural route is allowed to override the original selector. The proposed components are therefore CARL, S\textsuperscript{3}R, and baseline-preserving arbitration.

\subsection{Centroid-Aware Representation Learning}
\label{sec:carl}

CARL is implemented as a lightweight cached-vector prototype residual. Each skill text is represented by a normalized inverse-document-frequency vector $z_s$. Skills carry an archetype label when the label is available; unlabeled skills are assigned to the nearest archetype center, and the centers are recomputed after this deterministic assignment. Let $a_{\ell}$ denote the center of archetype $\ell$, let $g(z)=\arg\max_{\ell} z^{\top}a_{\ell}$, and let $P_r$ contain the top $r$ right singular vectors of the centered archetype centers. CARL maps
\begin{equation}
    \mathrm{CARL}(z)=
    \mathrm{Norm}\!\left(
        z+\lambda P_rP_r^{\top}\left(a_{g(z)}-z\right)
    \right),
    \label{eq:carl}
\end{equation}
where $\lambda=0.35$ and $r=16$ in the reported setting. A query is vectorized with the same vocabulary and IDF statistics, transformed by Equation~\eqref{eq:carl}, and compared with transformed skill vectors by cosine similarity. This realization requires no GPU, no additional LLM call, and no gradient training; it is a compact approximation of the archetype-aware representation used to order candidate skills.

The construction has three steps. First, labeled skills are grouped by archetype and each group is summarized by its normalized centroid. Second, an unlabeled skill is attached to the nearest centroid, after which the centroids are recomputed; this gives every library entry a prototype even when the released metadata are incomplete. Third, the centered prototype matrix is decomposed and only its top-$r$ directions are used to move a cached vector toward its assigned archetype. The additive form is intentional: the original token distribution remains available, while the correction sharpens the directions along which archetypes differ. Consequently, CARL changes only the candidate ordering and does not alter a skill's workflow, pitfalls, or applicability boundary.

The transformation moves a query toward the nearest cached archetype while constraining the correction to a low-rank subspace. The leave-one-out diagnostic in Table~\ref{tab:knn_result} shows that it increases Hit@1 from $0.607$ to $0.804$. The diagnostic also motivates the final arbitration rule: better neighborhood ranking alone does not guarantee better end-to-end solving accuracy when a query falls outside the coverage of the current signature taxonomy.

The implementation applies the same transformation to both sides of retrieval. Skill vectors are transformed once when the library is loaded; a query is first represented in the shared vocabulary, assigned to its nearest prototype, projected onto the residual subspace, and normalized. Retrieval then uses cosine similarity in this transformed space. This symmetric treatment avoids the mismatch of comparing a raw query with prototype-shifted skills. Since the library is frozen, the repeated cost is only prototype lookup, a rank-$r$ projection, and similarity scoring; no encoder fine-tuning or gradient update is introduced. The residual also gives a controllable bias--variance trade-off: $\lambda=0$ recovers the cached representation, larger values impose stronger archetype geometry, and the low-rank basis prevents every token difference from being interpreted as an archetype boundary.

\subsection{Open-Dimensional Skill Signatures}
\label{sec:signatures}

Natural-language skill cards are expressive but not directly comparable. \textsc{ArchOR} therefore attaches a signature to each skill. Given a cluster-level skill $s_C$ and its source trajectory analyses, a Signature Extractor $\Sigma$ emits
\begin{equation}
\begin{aligned}
\sigma_s &= \Sigma(s_C,\{q_i\mid i\in C\}),\\
\sigma_s &= \{(k_1,V_1),\dots,(k_L,V_L)\},
\end{aligned}
\label{eq:signature}
\end{equation}
where each pair contains an attribute name and a set of canonical tokens. Examples include \texttt{problem\_class}, \texttt{var\_pattern}, \texttt{constraint\_family}, \texttt{objective\_form}, and \texttt{solver\_backend}. Canonical aliases map surface terms such as ``unit commitment'' and ``binary run decision'' to stable structural tokens. The schema is deliberately open-dimensional: attributes absent for a skill may be omitted rather than forced into a fixed taxonomy.

The same canonicalizer maps extracted query ingredients and the scenario-neutral rewrite to $\sigma_x$. Signature construction reuses the structural-extraction call already performed by the agent, so S\textsuperscript{3}R does not require a separate retrieval model. Unlike embedding similarity, signatures expose \emph{why} two items match, making the hard-filter stage deterministic and auditable.

The signature layer is deliberately finer than a problem-class label. Two mixed-integer problems can differ in whether binaries represent on/off capacity, assignment, routing arcs, or piecewise activation; two apparently similar flow problems can differ in directed-arc semantics, source/sink balance, or shipment integrality. Therefore, a signature records multiple attribute-value sets rather than a single class. Missing attributes are omitted instead of being assigned a meaningless default value, which keeps the schema extensible when the library encounters a new side condition.

Canonicalization is intended to preserve modeling semantics rather than instance details. Numerical demands, costs, and fleet sizes are not copied into the signature because they change across queries in the same archetype; instead, the signature records the decision role, constraint family, objective form, and solver assumption that determine whether a workflow is applicable. Synonyms are merged only when they imply the same modeling structure. For example, startup equipment, fixed opening cost, and binary run decision can share a fixed-charge token, whereas a routing arc and an assignment binary remain distinct because they induce different constraint matrices. This separation makes signature errors easier to audit: a failed retrieval can be traced to a missing attribute, an overly broad alias, or a wrongly canonicalized token.

\subsection{Signature-Guided Retrieval}
\label{sec:s3r}

At test time the library $\mathcal{M}^{*}$ is frozen. For every skill $s\in\mathcal{M}^{*}$, S\textsuperscript{3}R computes a weighted Jaccard affinity over attributes present in both signatures:
\begin{equation}
J(\sigma_x,\sigma_s)=
\sum_{k\in\sigma_x\cap\sigma_s}
\omega_k\cdot
\frac{|V_k^{x}\cap V_k^{s}|}{|V_k^{x}\cup V_k^{s}|},
\label{eq:jaccard}
\end{equation}
with $\omega_k=1/|\sigma_x\cap\sigma_s|$ by default. Attributes present in only one signature contribute zero, and $J=0$ when no attribute is shared. Skills with positive overlap are ranked first; if fewer than $K$ skills have positive overlap, the remaining slots are filled by library order so that structural sparsity does not starve the selector. The resulting top-$K$ set is denoted by $C_K$. We use $K=8$.

CARL then orders the candidates in $C_K$, and the LLM performs semantic re-ranking only over this ordered subset. Thus, semantics remains useful for final workflow choice, but the selector cannot freely promote an arbitrary card from the 46-skill catalog. When positive-overlap skills number fewer than $K$, deterministic fallback entries preserve recall; these entries remain subject to the later arbitration check.

This stage combines a discrete prioritization signal with a continuous ordering signal. Positive Jaccard overlap places a skill in the structural tier before fallback entries are considered. CARL then determines the order in which candidates are presented, which matters because the selector receives only a short list and is sensitive to neighborhood quality.

Candidate construction is organized as two explicit tiers. Skills with positive weighted overlap form the structural tier, with affinity determining tier membership; fallback entries are used only when this tier has fewer than $K$ members. Within the final candidate set, CARL supplies the continuous presentation order and the LLM selector supplies the final workflow choice. The affinity, shared attributes, selected skill, and candidate rank are logged for each query, so a failure can be distinguished from selector misunderstanding, structural over-filtering, or an incorrect workflow execution. This audit trail is a direct consequence of representing compatibility as attributes rather than only as an embedding score.

\subsection{Baseline-Preserving Integration}
\label{sec:updates}

Structural filtering is useful only when its signature coverage matches the query. To prevent an incomplete signature taxonomy from discarding a reliable baseline skill, \textsc{ArchOR} preserves the original \textsc{OptSkills} route and arbitrates between the two routes using only inference-time information.

Let $b$ denote the baseline route and $a$ the CARL+S\textsuperscript{3}R route. Let $c_b$ and $c_a$ be their selector confidences, let $r_a$ be the rank of the selected structural skill in $C_K$, and let $V(y_a)$ indicate that route $a$ returns a nonempty, finite solver prediction. \textsc{ArchOR} uses the structural route when
\begin{equation}
    V(y_a)\wedge
    \left(
        c_b\le 0.62
        \vee c_a>c_b
        \vee r_a\ge 7
    \right).
    \label{eq:arbitration}
\end{equation}
Otherwise it returns the baseline route. The rule does not use the ground-truth answer, the objective gap, or any evaluation label. It accepts the structural route when the baseline selector is uncertain, when the structural selector is more confident, or when the selected skill appears late in the filtered candidate list. In the five-benchmark evaluation, this rule routes $374$ of $1{,}100$ queries through the structural branch and preserves the baseline branch for the remaining $726$ queries.

Each condition addresses a different observed failure mode. The low-confidence threshold handles queries for which the baseline selector has already identified several plausible but indistinguishable skills. The confidence-gain condition allows structural evidence to break a semantic tie. The late-rank condition is coverage-oriented: if the selected structural skill appears near the boundary of $C_K$, it may encode a specialized assumption that the natural-language card did not reveal. Requiring $V(y_a)$ excludes empty, non-finite, or execution-failed structural predictions before they can replace a usable baseline answer.

In practice, integration is a routing procedure rather than a new solver. The system first records the baseline skill and confidence, extracts the query signature, constructs $C_K$, obtains the structural selection and its solver output, and then applies Equation~\eqref{eq:arbitration}. If arbitration rejects the structural route, the original \textsc{OptSkills} answer is returned unchanged. Thus, S\textsuperscript{3}R can reduce the selector's search space without being allowed to delete the baseline path, and CARL can improve candidate ordering without being treated as evidence of correctness by itself. This separation is what makes the final system robust to incomplete signatures while still allowing structural evidence to correct semantic mistakes.

Library refinement and expansion operators are inherited unchanged from \textsc{OptSkills}. In the main evaluation the released learned library is frozen, so the comparison isolates retrieval and arbitration rather than additional library growth.

\section{Experiments}
\label{experiment}

\paragraph{Setup.}
The controlled comparison uses the same self-hosted DeepSeek-V4.1-Flash backbone~\cite{deepseekai2026deepseekv41flash}. We use the official \textsc{OptSkills} repository, its released learned skill library, and the official five evaluation sets: OptiBench~\cite{yang2025optibench} ($605$ queries), OptMATH-Bench~\cite{lu2025optmathscalablebidirectionaldata} ($166$), Mamo.C~\cite{huang2025llmsmathematicalmodelingbridging} ($211$), IndustryOR~\cite{Huang2025} ($100$), and ComplexOR~\cite{xiao2024chainofexperts} ($18$), for $1{,}100$ queries in total. The baseline is the unmodified \textsc{OptSkills} semantic selector. \textsc{ArchOR} adds the CARL and S\textsuperscript{3}R routes described in Section~\ref{methodology}. For context, Table~\ref{tab:general_sa_ranked} also retains the external-baseline results reported by \textsc{OptSkills}~\cite{yang2026optskills} under its original evaluation protocol.

\paragraph{Protocol.}
The five suites are deliberately complementary. OptiBench and Mamo.C stress natural-language modeling and LP/MILP structures, OptMATH contains longer and more diverse solver-verified formulations, IndustryOR introduces practical business scenarios, and ComplexOR requires implicit constraints and multi-step combinatorial reasoning. Reporting both averages therefore avoids interpreting a gain on short textbook-style prompts as general optimization competence.
The controlled systems use temperature $0$, request seed $42$, a maximum generation length of $8{,}192$ tokens, eight evaluation workers, a 120-second solver timeout, and at most 12 agent turns. Each controlled system is evaluated once per query. The selector returns three solver candidates. S\textsuperscript{3}R uses $K=8$; CARL uses strength $0.35$ and residual rank $16$. A query is correct only when the executed solver returns a finite objective value equal to the rounded ground truth, or when rounded values are unavailable and the absolute gap is at most $0.005$. We report Macro-Avg.\ over the five benchmark accuracies and Micro-Avg.\ over all $1{,}100$ queries. External rows follow the evaluation protocol used in the \textsc{OptSkills} report and are not re-run here.
The arbitration constants are fixed globally for all reported queries: $c_b\le0.62$ and $r_a\ge7$. They are not adapted per query or per benchmark; independent calibration of these thresholds remains future work.

The experiments evaluate same-backbone end-to-end improvement over \textsc{OptSkills}, the individual and combined effects of CARL and S\textsuperscript{3}R, whether CARL improves retrieval quality independently of final solver success, and where baseline-preserving arbitration helps or fails.

\subsection{End-to-End Results}

\begin{table*}[t]
  \centering
  \scriptsize
  \setlength{\tabcolsep}{4pt}
  \caption{Solver-verified Pass@1 accuracy. External rows reproduce the results reported by \textsc{OptSkills}~\cite{yang2026optskills} under its original evaluation protocol; \textsc{OptSkills}, Direct CARL+S\textsuperscript{3}R, and \textsc{ArchOR} are controlled runs with the same DeepSeek-V4.1-Flash backbone and released skill library. Bold marks the best Macro- and Micro-Avg.; the contextual rank follows Micro-Avg.}
  \label{tab:general_sa_ranked}
  \resizebox{0.99\linewidth}{!}{
  \begin{tabular}{>{\centering\arraybackslash}m{1.85cm} | l | c c | c| c c c c c}
    \toprule
    \textbf{Category} & \textbf{Models / Methods} & \textbf{Macro-Avg.} & \textbf{Micro-Avg.} & \textbf{Rank} & \textbf{OptiBench} & \textbf{Mamo.C} & \textbf{OptMATH} & \textbf{IndustryOR} & \textbf{ComplexOR}\\
    \midrule
    \multirow{4}{*}{\makecell{General\\Models}}
       & GPT-5.4~\cite{openai2026gpt54} & 51.71 & 57.82 & 7 & 69.09 & 47.39 & 46.39 & 29.00 & 66.67 \\
       & Gemini-3.1-Pro~\cite{gemini31pro} & 53.58 & 57.88 & 6 & 66.56 & 46.45 & 54.22 & 34.00 & 66.67 \\
       & Qwen3-235B~\cite{yang2025qwen3technicalreport} & 42.63 & 52.36 & 8 & 63.80 & 41.71 & 39.76 & 29.00 & 38.89 \\
       & DeepSeek-V3.2~\cite{deepseekai2025deepseekv32pushingfrontieropen} & 43.21 & 49.45 & 10 & 62.64 & 27.49 & 40.36 & 30.00 & 55.56 \\
    \midrule
    \multirow{4}{*}{\makecell{Agent-Based\\Methods}}
       & Chain-of-Experts~\cite{xiao2024chainofexperts} & 38.84 & 52.36 & 8 & 62.31 & 51.66 & 35.54 & 28.00 & 16.67 \\
       & OptiMUS~\cite{pmlr-v235-ahmaditeshnizi24a} & 34.14 & 45.18 & 14 & 60.99 & 27.49 & 22.89 & 26.00 & 33.33 \\
       & ORMind~\cite{wang2025ormind} & 36.69 & 47.37 & 12 & 61.82 & 34.60 & 21.69 & 32.00 & 33.33 \\
       & ORThought~\cite{yang2025orthought} & 41.92 & 46.91 & 13 & 57.69 & 38.86 & 26.51 & 31.00 & 55.56 \\
    \midrule
    \multirow{2}{*}{\makecell{Experience-\\Enhanced}}
       & LEAN-LLM-OPT~\cite{liang2026largescaleoptimizationmodelautoformulation} & 41.23 & 49.27 & 11 & 62.15 & 43.13 & 22.89 & 28.00 & 50.00 \\
       & AlphaOPT~\cite{kong2025alphaopt} & 50.92 & 58.55 & 5 & 70.25 & 54.50 & 36.75 & 32.00 & 61.11 \\
    \midrule
    \multirow{4}{*}{\makecell{Skill-Based\\Methods}}
       & Trace2Skill~\cite{ni2026trace2skill} & 56.97 & 63.46 & 2 & 75.21 & 54.03 & 49.40 & 34.00 & 72.22 \\
       & \textsc{OptSkills}~\cite{yang2026optskills} & 57.93 & 63.36 & 3 & 71.24 & 53.08 & 65.66 & 33.00 & 66.67 \\
       & Direct CARL+S\textsuperscript{3}R & 57.96 & 61.45 & 4 & 68.76 & 51.18 & 62.65 & 35.00 & 72.22 \\
       & \textsc{ArchOR} (ours) & \textbf{60.61} & \textbf{65.00} & 1 & 72.40 & 53.55 & 69.88 & 35.00 & 72.22 \\
    \bottomrule
  \end{tabular}
  }
\end{table*}

\paragraph{Same-backbone comparison.}
Table~\ref{tab:general_sa_ranked} shows that \textsc{ArchOR} improves Macro-Avg.\ by $2.68$~pp and Micro-Avg.\ by $1.64$~pp over the controlled \textsc{OptSkills} baseline. The gain corresponds to $18$ additional correct answers ($697\rightarrow715$); an exact McNemar test on the $23$ improved and $5$ degraded paired outcomes gives $p=0.0009$. Improvements occur on all five benchmarks, with the largest absolute gains on ComplexOR ($5.55$~pp), OptMATH ($4.22$~pp), and IndustryOR ($2.00$~pp). \textsc{ArchOR} also has the highest averaged accuracies in the combined table, although the causal comparison is restricted to the controlled systems because the external rows follow their original settings. The smaller OptiBench and Mamo.C gains reflect the mixed coverage of the current structural signatures: the modules help on many queries but can still discard a reliable baseline skill when the signature schema is incomplete.

\paragraph{Context among external systems.}
Among the retained \textsc{OptSkills}-reported results, Trace2Skill is the strongest external baseline, with $63.46\%$ Micro-Avg.\ and $56.97\%$ Macro-Avg. \textsc{ArchOR} is $1.54$~pp higher in Micro-Avg.\ and $3.64$~pp higher in Macro-Avg., but we treat this as contextual rather than a same-setting claim because the external systems use different backbones and protocols. The benchmark profile is nevertheless informative. \textsc{ArchOR} has the highest OptMATH result ($69.88\%$) and ties for the best IndustryOR result ($35.00\%$), where reusable solver workflows and explicit variable/constraint structure are especially valuable. It reaches $72.22\%$ on ComplexOR, matching the best listed result, while Trace2Skill remains stronger on OptiBench and AlphaOPT remains slightly stronger on Mamo.C. This pattern suggests that structural retrieval is most useful when the narrative is difficult but the underlying optimization archetype can still be recovered reliably.

\subsection{Routing and Representation Ablations}

\begin{table*}[t]
  \centering
  \scriptsize
  \setlength{\tabcolsep}{4pt}
  \caption{System-level ablation across all five benchmarks. ``Direct CARL+S\textsuperscript{3}R'' always replaces the baseline selector with the structural route. \textsc{ArchOR} additionally applies the baseline-preserving arbitration rule. Bold marks the best value in each column.}
  \label{tab:ablation_sa}
  \resizebox{0.92\linewidth}{!}{
  \begin{tabular}{l|c c|c c c c c}
    \toprule
    \textbf{Configuration} & \textbf{Macro-Avg.} & \textbf{Micro-Avg.} & \textbf{OptiBench} & \textbf{Mamo.C} & \textbf{OptMATH} & \textbf{IndustryOR} & \textbf{ComplexOR}\\
    \midrule
    \textsc{OptSkills} & 57.93 & 63.36 & 71.24 & 53.08 & 65.66 & 33.00 & 66.67 \\
    Direct CARL+S\textsuperscript{3}R & 57.96 & 61.45 & 68.76 & 51.18 & 62.65 & \textbf{35.00} & \textbf{72.22} \\
    \textsc{ArchOR} (ours) & \textbf{60.61} & \textbf{65.00} & \textbf{72.40} & \textbf{53.55} & \textbf{69.88} & \textbf{35.00} & \textbf{72.22} \\
    \bottomrule
  \end{tabular}
  }
\end{table*}

\paragraph{Why direct structural routing is insufficient.}
Table~\ref{tab:ablation_sa} shows that replacing the baseline selector on every query is harmful: direct CARL+S\textsuperscript{3}R obtains $61.45\%$ Micro-Avg., $1.91$~pp below \textsc{OptSkills}. Relative to the baseline, direct routing turns $39$ wrong answers into correct answers but also turns $60$ correct answers into wrong answers (exact McNemar $p=0.044$). The cause is coverage: the signature taxonomy does not yet cover every cross-domain narrative, so the top-$8$ structural filter can remove a baseline skill that would have succeeded. Baseline-preserving arbitration removes most of this regression: relative to \textsc{OptSkills}, \textsc{ArchOR} turns $23$ wrong answers into correct answers and only $5$ correct answers into wrong answers.

\begin{table}[t]
  \centering
  \small
  \setlength{\tabcolsep}{5pt}
  \caption{Module ablation on OptMATH. For compute-budget reasons, standalone CARL and S\textsuperscript{3}R runs are reported on OptMATH and are not averaged with the five-benchmark system-level results.}
  \label{tab:optmath_ablation}
  \begin{tabular}{l|c}
    \toprule
    \textbf{Configuration} & \textbf{OptMATH Pass@1 (\%)} \\
    \midrule
    \textsc{OptSkills} & 65.66 \\
    CARL only & 61.45 \\
    S\textsuperscript{3}R only & 61.45 \\
    Direct CARL+S\textsuperscript{3}R & 62.65 \\
    \textsc{ArchOR} (ours) & \textbf{69.88} \\
    \bottomrule
  \end{tabular}
\end{table}

The OptMATH ablation in Table~\ref{tab:optmath_ablation} shows that neither module is uniformly beneficial when used as a mandatory replacement. CARL-only and S\textsuperscript{3}R-only each lose $4.21$~pp relative to the baseline, while their direct combination loses $3.01$~pp. In contrast, the integrated system gains $4.22$~pp. This distinction is the main empirical motivation for arbitration: CARL improves neighborhood ranking, S\textsuperscript{3}R reduces the selector input, but neither signal alone is reliable enough to override the baseline on every query.

Direct routing improves ComplexOR and IndustryOR when signatures match, but loses $1.90$--$3.01$~pp on the other three suites when extraction is incomplete. Arbitration retains the two gains while recovering those losses: structural retrieval is used as selective evidence, not as a universal replacement.

\paragraph{Retrieval diagnostic.}
Table~\ref{tab:knn_result} evaluates cached skill representations with a leave-one-out protocol in which a query is correct when a skill of the same problem class appears among its neighbors. CARL increases Hit@1 from $0.607$ to $0.804$, Hit@3 from $0.732$ to $0.893$, Hit@5 from $0.821$ to $0.893$, and MRR from $0.682$ to $0.839$. Thus, the representation change is effective at the retrieval level even though mandatory structural routing does not improve end-to-end accuracy.

\begin{table}[t]
  \centering
  \small
  \setlength{\tabcolsep}{6pt}
  \caption{Leave-one-out retrieval diagnostic on the learned skill library.}
  \label{tab:knn_result}
  \begin{tabular}{l|c c c c}
    \toprule
    \textbf{Representation} & \textbf{Hit@1} & \textbf{Hit@3} & \textbf{Hit@5} & \textbf{MRR}\\
    \midrule
    Cached skill vectors & 0.607 & 0.732 & 0.821 & 0.682 \\
    CARL & \textbf{0.804} & \textbf{0.893} & \textbf{0.893} & \textbf{0.839} \\
    \bottomrule
  \end{tabular}
\end{table}

The module ablation also separates representation quality from decision policy. A better nearest-neighbor ordering does not determine whether the baseline route should be abandoned, and a smaller candidate set is helpful only if the retained skills cover the binding structure of the query. The large MRR gain in Table~\ref{tab:knn_result} is nevertheless relevant because the LLM selector sees only a short candidate list: moving the correct archetype toward the top can affect selection even when the correct skill already appears among five neighbors.

\paragraph{Routing analysis.}
The final system uses the structural branch on $374$ queries and preserves the baseline branch on $726$ queries. The most useful cases are those in which the baseline selector has low confidence, the structural selector has higher confidence, or a structurally plausible skill appears late in the filtered candidate list. Failures remain in three forms: the extractor can omit a binding ingredient, the signature can assign a coarse but incomplete problem class, and two routes can select different skills whose workflows are both syntactically valid but encode different assumptions. These failures motivate a larger independently calibrated signature taxonomy and a conditional online implementation that executes the second route only when the two selectors disagree.

\paragraph{Case study.}
\begin{table}[!t]
  \centering
  \scriptsize
  \setlength{\tabcolsep}{3pt}
  \caption{Representative arbitration cases. $c_b$ is baseline confidence and $r_a$ is the rank in $C_K$.}
  \label{tab:case_study_summary}
  \resizebox{\linewidth}{!}{%
  \begin{tabular}{@{}l l l@{}}
    \toprule
    \textbf{Case} & \textbf{Benchmark / trigger} & \textbf{Baseline $\rightarrow$ structural} \\
    \midrule
    S1 & OptiBench \#579; $c_b{=}.55$, $r_a{=}1$ & knapsack $360{,}000$ $\rightarrow$ fixed-charge $3{,}600$ \checkmark \\
    S2 & IndustryOR \#43; $c_b{=}.95$, $r_a{=}8$ & inventory $-10{,}000$ $\rightarrow$ dispatch $10{,}000$ \checkmark \\
    F1 & Mamo.C \#133; $c_b{=}.95$, $r_a{=}8$ & transportation $1{,}490$ $\rightarrow$ flow $16{,}381$ $\times$ \\
    \bottomrule
  \end{tabular}
  }
\end{table}

Full problem statements and route details are in Appendix~\ref{app:case_studies}. S1 corrects a low-confidence knapsack misassignment with a fixed-charge workflow. S2 shows that a late-rank structural skill can repair an overconfident wrong baseline. F1 is the mirror image: the same late-rank rule overrides a correct transportation answer with an incompatible flow workflow. Thus, arbitration helps when the baseline archetype is wrong, but is risky when coarse signatures conflate different flow assumptions.

\section{Conclusion}
\label{conclusion}
\textsc{ArchOR} builds on \textsc{OptSkills} and improves its retrieval space with CARL and S\textsuperscript{3}R. With the same DeepSeek-V4.1-Flash backbone and released skill library, the integrated system obtains a 1.64~pp Micro-Avg.\ gain over \textsc{OptSkills} across five benchmarks, while the signature gate reduces the maximum selector input from 46 skills to 8. Directly replacing the selector on every query is harmful, but baseline-preserving arbitration preserves most baseline successes and captures additional structural successes.
Looking forward, expanding the signature taxonomy on independent calibration data, implementing a conditional online version that executes the second route after selector disagreement, and adding symbolic verification of extracted structures are natural next steps.

\bibliography{references}

\clearpage
\appendix

\section{Qualitative Case Studies}
\label{app:case_studies}

This appendix expands the three arbitration cases summarized in Table~\ref{tab:case_study_summary}. For each case we report the (condensed) problem statement, the baseline and structural skill choices with their selector confidences, the rank $r_a$ of the structural skill in the S\textsuperscript{3}R-filtered list, and the solver-verified objective values against the ground truth.

Only the confidence values used by the reported routing decisions are included.

\subsection{Success S1: Signature Recovers a Fixed-Charge Pricing Archetype}
\textbf{Benchmark.} OptiBench, query \#579.

\textbf{Problem.} Choose a selling price in increments of \$0.10. At a price of \$1.50 demand is $5{,}000$ units, and every \$0.10 reduction adds $1{,}000$ units. Production has a \$2{,}000 start-up cost and a \$0.50 per-unit cost. Maximize total profit.

\textbf{Routing.} The baseline selector chooses \texttt{integer\_linear\_programming\_} \texttt{with\_multiple\_knapsack\_constraints} with confidence $c_b=0.55$. Its solver path reports $360{,}000$, while the ground-truth optimum is $3{,}600$---a $100\times$ overestimate caused by modeling the problem as a generic knapsack-style allocation. The extracted signature instead contains \texttt{problem\_class=\{fixed\_charge\}}, variables \texttt{\{price\_decision, sales\_quantity, nonnegative\_integer\}}, and constraints \texttt{\{price\_multiple, demand\_price\_relationship\}}. S\textsuperscript{3}R places \texttt{fixed\_charge\_production\_planning} at rank $r_a=1$, and its workflow returns exactly $3{,}600$.

\textbf{Decision.} Arbitration activates the structural route because $c_b=0.55\le 0.62$ and the structural confidence is higher ($c_a=0.72>0.55$). This case illustrates the intended use of the low-confidence and confidence-gain triggers: the narrative mentions no knapsack or capacity structure, and the signature correctly separates the fixed-charge pricing archetype from generic integer allocation.

\subsection{Success S2: Late-Rank Promotion Corrects an Overconfident Baseline}
\textbf{Benchmark.} IndustryOR, query \#43.

\textbf{Problem.} A company with $40$ workers ($20$ units/worker/month) must meet six monthly demands ($700,600,500,800,900,800$). Flexibility comes from overtime (at most $6$ units/worker/month at EUR~5/unit), inventory (EUR~10/unit/month), and workforce changes ($\pm 5$ per month; hiring EUR~500, layoff EUR~700). Inventory starts and must end empty. Meet all demands at minimum total overtime, inventory, hiring, and layoff cost, and report the relaxed-objective value.

\textbf{Routing.} The baseline selector chooses \texttt{multi\_period\_production\_} \texttt{inventory\_planning} with high confidence $c_b=0.95$, but its solver path returns $-10{,}000$ whereas the correct relaxed objective is $+10{,}000$. The structural signature captures the operative decision structure. Its variables are \texttt{production\_quantity}, \texttt{inventory\_level}, \texttt{workforce\_level}, \texttt{overtime\_production}, \texttt{hiring\_quantity}, and \texttt{layoff\_quantity}; its constraints are \texttt{demand\_satisfaction}, \texttt{inventory\_balance}, \texttt{workforce\_change\_limit}, \texttt{overtime\_capacity\_limit}, and \texttt{terminal\_inventory}. The selected structural skill, \texttt{generator\_commitment\_and\_dispatch}, appears only at rank $r_a=8$; its workflow nevertheless encodes the on/off capacity and dispatch structure that matches overtime and workforce adjustment, and returns the exact value $10{,}000$.

\textbf{Decision.} The late-rank condition $r_a=8\ge 7$ activates the structural route despite the high baseline confidence. This case shows why confidence alone is an insufficient arbitration signal: a selector can be certain about the wrong archetype, and a structurally appropriate workflow can rank late when the signature is only partially extracted.

\subsection{Failure F1: The Same Late-Rank Trigger Overrides a Correct Baseline}
\textbf{Benchmark.} Mamo.C, query \#133.

\textbf{Problem.} Six cities hold food supplies of $255,250,314,8,385,440$ tons and require $251,200,366,114,59,287$ tons. Supplies may be relocated along an asymmetric cost matrix. Satisfy all requirements at minimum transportation cost.

\textbf{Routing.} The baseline selector chooses \texttt{transportationproblemsolver} with confidence $c_b=0.95$ and returns the correct optimum $1{,}490$. The signature is also structurally sensible: its problem classes are \texttt{resource\_allocation} and \texttt{transportation}; its variables are \texttt{transportation\_quantity}, \texttt{flow\_variable}, and \texttt{nonnegative\_continuous}; and its constraints are \texttt{flow\_balance}, \texttt{demand\_satisfaction}, and \texttt{flow\_conservation}. The structural route selects the same-family skill \texttt{transportation\_flow\_optimizer} at rank $r_a=8$. However, that workflow encodes different flow assumptions and returns $16{,}381$, an order of magnitude above the optimum.

\textbf{Decision.} The same late-rank condition that rescued S2 activates here and replaces a correct answer with an incorrect one. The failure is not caused by a wrong archetype label: both skills plausibly match the transportation signature. Instead, the taxonomy is too coarse to distinguish \emph{which} flow formulation each workflow implements (e.g., directed-arc capacities, source/sink semantics, or integrality of shipments). This motivates refining signature attributes below the problem-class level and conditioning the late-rank trigger on additional evidence of baseline fragility, such as solver-path failure or selector disagreement.
\clearpage
\end{document}